\documentclass[
]{ceurart}

\usepackage{listings}
\usepackage[]{mdframed}

\begin{document}

\copyrightyear{2023}
\copyrightclause{Copyright for this paper by its authors.
  Use permitted under Creative Commons License Attribution 4.0
  International (CC BY 4.0).}

\conference{Scholarly-QALD-23: Scholarly QALD Challenge at The 22nd International Semantic Web Conference (ISWC 2023), November 6 – 10, 2023, Athens, Greece.}

\title{Leveraging LLMs in Scholarly Knowledge Graph Question Answering}

\tnotemark[1]

\author[1]{Tilahun Abedissa Taffa}[%
orcid=0000-0002-2476-8335,
email=tilahun.taffa@uni-hamburg.de,
] 
\cormark[1]
\address[1]{Semantic Systems, Universität Hamburg, Vogt-Kölln-Straße 30, 22527 Hamburg, Germany}
\author[2]{Ricardo Usbeck}[%
orcid=0000-0002-0191-7211,
email= ricardo.usbeck@leuphana.de,
url=https://www.leuphana.de/institute/iis/personen/ricardo-usbeck.html,
]
\address[2]{Leuphana Universität Lüneburg, Universitätsallee 1, C 4.314, 21335 Lüneburg, Germany}

\cortext[1]{Corresponding author.}

\begin{abstract}
  This paper presents a scholarly Knowledge Graph Question Answering (KGQA) that answers bibliographic natural language questions by leveraging a large language model (LLM) in a few-shot manner. The model initially identifies the top-n similar training questions related to a given test question via a BERT-based sentence encoder and retrieves their corresponding SPARQL. Using the top-n similar question-SPARQL pairs as an example and the test question creates a prompt. Then pass the prompt to the LLM and generate a SPARQL. Finally, runs the SPARQL against the underlying KG - ORKG (Open Research KG) endpoint and returns an answer. Our system achieves an F1 score of 99.0\%, on SciQA - one of the Scholarly-QALD-23 challenge benchmarks.
  
\end{abstract}


\maketitle

\section{Introduction}\label{intro}

Scholarly \textbf{K}nowledge \textbf{G}raph \textbf{Q}uestion \textbf{A}nswering (KGQA) models answer machine or human-generated scholarly natural language questions over KGs that contain bibliographic metadata information~\cite{auer2020improving, jaradeh2020question}. The approaches used in the existing Scholarly KGQA models fall into two categories. The first type is a retriever-reasoner framework, which involves retrieving relevant sub-graphs and then using reasoning to extract entities as answers~\cite{{wang2021literatureqa}}. The second type is a semantic parsing-based framework, which focuses on transforming questions into executable logical expressions like SQL or SPARQL that can be used to obtain the answer(s) by querying the underlying KG~\cite{banerjee2023dblp}. %SPARQL (SPARQL Protocol and RDF Query Language) is a query language designed for querying RDF (Resource Description Framework) data\footnote{https://www.w3.org/TR/sparql11-query/}. 
However, both the retriever-reasoner and semantic parsing approaches need a large amount of training data to create a robust KGQA model. Specifically, the scarcity of scholarly KGQA data sets makes the task more challenging than other general KGQA. Hence, one of the possible solutions is exploring the power of Large Language Models (LLMs) in a zero or few-shot manner.

\begin{figure}[htb!]
\label{fig:sample-query}
  \centering
  \includegraphics[width=\linewidth]{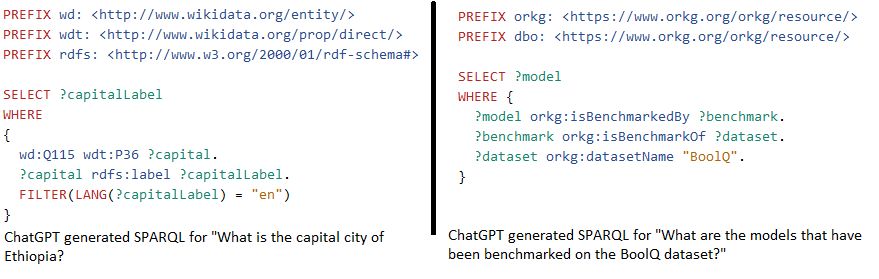}
  \caption{SPARQL queries generated by ChatGPT 3.5 in a zero-shot manner for querying Wikidata (left) and ORKG (right).}
\end{figure}

LLMs are trained on a large amount of textual data for tackling human language understanding and generating tasks~\cite{NEURIPS2020_1457c0d6, touvron2023llama}. LLMs enormous amount (counted in billions) of parameters and adaptive capability in AI (Artificial Intelligence) applications have contributed to the creation of robust QA models~\cite{banerjee2023gett, chen-2023-large, kamalloo-etal-2023-evaluating, ziems-etal-2023-large}. Besides that, the recent advancement in prompt engineering\footnote{The process of designing prompts that help LLMs to perform a task}, empowers everyone to get the most out of LLMs~\cite{sorensen-etal-2022-information}. For instance, few-shot LLM prompting, the method of instructing the LLM with a minimal set of examples or context to perform a specific language generation task, generally yields more accurate and contextually relevant results~\cite{chen-2023-large}. Thus, by providing a few relevant question-SPARQL pairs, models like GPT-3~\cite{NEURIPS2020_1457c0d6} and its successors, can generalize and generate correct SPARQL queries to query encyclopedic KGs like Wikidata. For example, as shown in Figure~\ref{fig:sample-query} (left), ChatGPT 3.5\footnote{\url{https://openai.com/blog/chatgpt}} generates a correct query for the question ``What is the capital city of Ethiopia?" in zero-shot mode. Unlike that, for the scholarly question taken from SciQA test set ``What are the models that have been benchmarked on the BoolQ dataset?", even though the SPARQL generated in zero-shot has no syntactic errors (see Figure~\ref{fig:sample-query} right), it does not yield the right answer when run against the ORKG-dump SPARQL endpoint. This is because ChatGPT does not know the schema of ORKG. One way of addressing this knowledge gap in LLMs is using a few-shot approach.

Therefore, in this work, we harness the capabilities of LLMs to transform natural language questions into SPARQL queries. %The SPARQL generation task is very hard and time-consuming, especially for users who are not proficient in SPARQL or have limited knowledge of the underlying data schema. This is where the motivation for leveraging LLMs in SPARQL generation within Scholarly KGQA becomes evident. Generally, integrating LLMs for SPARQL generation in Scholarly KGQA is an option to democratize scholarly knowledge access. The primary motivation behind this paper is to contribute to the ongoing efforts to advance KGQA capabilities within the scholarly domain. 
We believe that successfully addressing the SciQA challenge at the Scholarly QALD Challenge\footnote{\url{https://kgqa.github.io/scholarly-QALD-challenge/2023/}} underway at ISWC 2023\footnote{\url{https://iswc2023.semanticweb.org}} allows us to contribute to enhancing access to and utilization of scholarly knowledge.

The contributions of our work are:
\begin{itemize}
    \item Leveraging LLMs for SPARQL query generation in a few-shot manner;
    %\item Avails a running Scholarly KGQA system
    \item Identifying similar questions using a BERT-based model~\cite{KOCAMAN2021100058};
    \item Developing a Scholarly KGQA model that ranked second in the SciQA Challenge leaderboard;
    \item Evaluating the impact of single shot and few-shot prompting for SPARQL generation performance.
\end{itemize}

Our source code can be found at \url{https://github.com/huntila/scholarly-kgqa}.
%The remaining part of this paper is organized as follows: Section~\ref{sota} depicts state-of-the-art challenges, Section~\ref{approach} describes our model, and Section~\ref{eval} contains the evaluation results and review of the test data. Furthermore, Section~\ref{eval} contains the SciQA data set description, evaluation results, discussion, and error analysis; and Section~\ref{summary} summarizes.

\section{Related Works}\label{sota}
\subsection{Scholarly KGQA}\label{related-work-schkgqa}
Pipeline-based, semantic parsing-based scholarly KGQA systems first identify the entities and relationships in the given question; and map those entities and relationships to their respective identifier in the KG. Next, formulate a query, e.g. SPARQL, and finally execute the query against the underlying KG and return an answer~\cite{ZHANG20231}. JarvisQA~\cite{jaradeh2020question}, create triples from the tables, then convert the triples to text, and extract an answer using a Bidirectional Encoder Representations from Transformers (BERT)~\cite{bertDevlinCLT19} based answer retriever. JarvisQA only operates on tabular data. Besides, the performance of the model is highly dependent on the correctness of the transformation of the table entries to triples and triples to text. Instead of using triples to text transformer and retriever, DBLP-QuAD~\cite{banerjee2023dblp}, parse questions to a SPARQL by fine-tuning a Text-to-Text Transfer Transformer (T5)~\cite{t5raffel} model. To use the DBLP-QuAD parser for a new scholarly KG with a different schema, fine-tuning requires a large amount of training data and an entity linker. Unlike JarvisQA, our model translates questions into SPARQL without the need for triple-to-text conversion. Additionally, our approach differs from DBLP-QuAD in that it employs an LLM to generate SPARQL with very few examples, which avoids the LLM pre-training.

\subsection{Few-Shot LLM Prompting}\label{few-shot-prompting}

In this work, we use Vicuna-13B\footnote{\url{https://lmsys.org/blog/2023-03-30-vicuna/}} - an open-source LLM, a descendent of LLaMA~\cite{touvron2023llama} fine-tuned using user-shared conversations collected from ShareGPT\footnote{\url{https://sharegpt.com}}. LLMs have demonstrated their remarkable ability to understand and generate natural language text in zero-shot and few-shot manner~\cite{NEURIPS2020_1457c0d6, baek-etal-2023-knowledge, chen-2023-large}. In few-shot prompting, the LLM is given the task description in natural language such as `generate SPARQL for the given questions' with few examples, then prompted to accomplish the task without any fine-tuning~\cite{chen-2023-large, sorensen-etal-2022-information}. So, in our work, we use few-shot prompting for question's SPARQL generation.

\begin{table}[htb!]
  \caption{SciQA dataset size.}
  \label{tab:sciqa_size}
  \begin{tabular}{llll}
    \toprule
     Data & train & dev & test \\
    \midrule
    Size & 1795 & 200 & 200 \\
    \bottomrule
  \end{tabular}
\end{table}

\section{The Scholarly QALD Challenge}\label{sciqa-dataset}

To foster standard evaluation of KGQA models, there have been a series of QALD (Question Answering over Linked Data) challenges since 2011~\cite{usbeckqald}. The datasets released in the past QALD challenges are based on generic KGs like Wikidata. Unlike that, the Scholarly QALD Challenge organized at the ISWC 2023 comes up with two new Scholarly QA data sets, namely SciQA (Scientific QA)~\cite{auer2023sciqa} and DBLP-QUAD~\cite{banerjee2023dblp}; and provides Codalab~\cite{codalab_competitions_JMLR} as a competition platform.

The SciQA benchmark data set - the challenge we participated in - is created following manual and template-based automatic generation methods. That is, first, 100 questions are created manually, afterward, from the manual questions curated eight questions and query templates. The manually created questions and queries underwent rigorous peer review by the authors and domain experts for correctness and relevance. Then, generate an additional three question and query templates from the eight question and query templates using  GPT-3~\cite{NEURIPS2020_1457c0d6}. Finally, 2465 questions are auto-generated by replacing entities and relations in the templates. All questions focus on Computer Science research works~\cite{auer2023sciqa}. Table~\ref{tab:sciqa_size} shows the train, dev, and test question split size of the SciQA dataset\footnote{\url{https://github.com/debayan/scholarly-QALD-challenge/tree/main/2023/datasets/sciqa/SciQA-dataset}}.

\section{The Scholarly KGQA Model}\label{approach}

\begin{figure}[htb!]
  \centering
  \includegraphics[width=\linewidth]{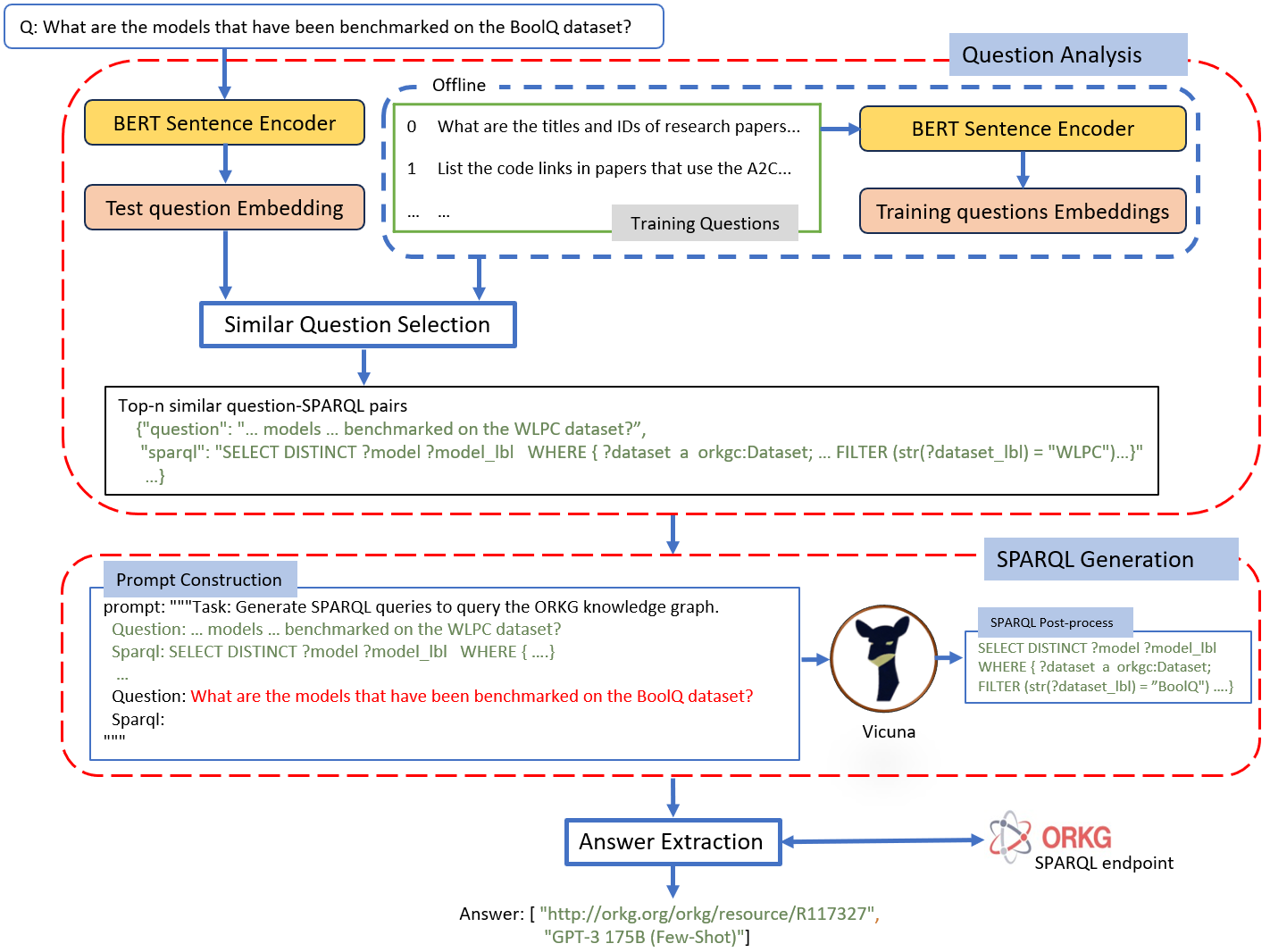}
  \caption{The Scholarly KGQA model.}
  \label{fig:sciqamodel}
\end{figure}

As shown in Figure~\ref{fig:sciqamodel} for a given question $Q$, our scholarly KGQA model encodes the training questions and $Q$. Then, it identifies a set of similar questions to $Q$ from the training set and constructs a prompt. Subsequently, the system generates $Q$'s SPARQL query by prompting the LLM and finally provides an answer $A$ by running the SPARQL against the ORKG SPARQL endpoint\footnote{\url{https://ltdemos.informatik.uni-hamburg.de/orkg/sparql}}. In the following, we explain in detail the three phases: question analysis, SPARQL generation, and answer extraction.

\subsection{Question Analysis}
This phase aims to identify the top-n questions from the training set similar to test question $Q$ and to fetch their respective SPARQL. As a result, the question and SPARQL pairs are used in the prompt formulation of the query generation. Therefore, the question analyzer first generates the question embedding score of each question in the training set offline using the BERT-based sentence encoder~\cite{KOCAMAN2021100058}. Besides, encodes the input test question using the same sentence encoder. Then compute the similarity score based on cosine similarity for each test question Q, rank the training questions based on their similarity score with $Q$, and select the top 5 questions along with their SPARQL.

 \subsection{Query Generation}\label{query_generation}

 This component addresses the challenge of extracting the correct entities and relationships from $Q$ and mapping them to the correct SPARQL query. Therefore, the query generation component constructs a prompt using the prompt template shown ~\ref{prompt_template}. In the prompt, an example is created by concatenating top-n (n=1,3,5) similar questions with their respective SPARQL queries. In the case of one shot, the example variable contains only one top-ranked question SPARQL pair. Whereas, in the three or five shot the example variable contains three or five top-ranked questions SPARQL pairs respectively. Before including the SPARQL queries in the prompt, the query generator removes special characters such as new lines, escaping characters, and extra blank spaces. 
 \begin{mdframed}\label{prompt_template}
\begin{verbatim}
  example = "Question: {sim_question} \n Sparql: {sparql}..."
  prompt: """ Task: Generate SPARQL queries to query the ORKG.
  Instruction: If you cannot generate a SPARQL query based 
              on the provided examples, explain the reason.
              {example}
  Question: {test question}
  Sparql:
  Note:  Output only the SPARQL query. """
\end{verbatim}
\end{mdframed}
 Then the SPARQL generator sub-component runs the prompt against our own Vicuna instance\footnote{\url{https://lmsys.org/blog/2023-03-30-vicuna/}}; and the output of the LLM is returned as the SPARQL of the test question $Q$.

\subsection{Answer Extraction}

The answer extractor receives the generated SPARQL queries and cleans new lines, escaping characters, and extra blank spaces, again. Finally runs the query against the underlying KG, in our case the ORKG endpoint\footnote{\url{https://ltdemos.informatik.uni-hamburg.de/orkg/sparql}}, and returns the result as an answer.

\section{Evaluation}\label{eval}

\subsection{Results and Discussion}\label{result-discussion}

The intuitive idea behind our model design is that we can get the best out of the LLM by providing prompts that contain similar question-SPARQL pairs and a test question. This makes the LLM self-learn and generates correct SPARQL based on examples. Hence, as shown in Table~\ref{tab:eval_results} and the Codalab SciQA Challenge leaderboard\footnote{\url{https://codalab.lisn.upsaclay.fr/competitions/public\_submissions/14759}} our scholarly KGQA model achieved a near-perfect F1 score of 0.99 using the top-3 similar questions. Queries generated using one-shot recorded the lowest F1 score of 0.96. In the one-shot setting, the LLM receives only the top-most similar question SPARQL pair example. %With this, the generated SPARQLs give a comparable F1 score. 
When using the top-5 similar questions, the F1 score reaches 0.989. The top-5 performance of our model is lower than in the top-3 model, because, as the number of question-SPARQL examples increases, it is likely that the probability of including training questions with less similarity to the test question. Thus, the inclusion of dissimilar question-SPARQL pairs confuses the LLM and generates queries that do not give the correct answer.

\begin{table}[htb!]
  \caption{F1 score of our model on SPARQLs generated using one, three, and five-shot methods on dev and test data.}
  \label{tab:eval_results}
  \begin{tabular}{lll}
    \toprule
    Data & Shot & F1 Score\\
    \midrule
    & One-shot & 0.960  \\
    test & Three-shot & \textbf{0.990} \\
    & Five-shot & 0.989 \\
    \midrule
    & One-shot & 0.899  \\
    dev & Three-shot & \textbf{0.974} \\
    & Five-shot & 0.973 \\
    \bottomrule
  \end{tabular}
\end{table}

Apart from that, the performance of our model is almost near to 1. The contributing factors are bifold. First, the test questions do not contain additional questions that are generated by those templates used to generate the training set. Thus, the LLM easily memorizes the entities and relations in the data set. Besides, the SPARQL queries in the data set use the literal indicating prefixes \textit{orkgc}, \textit{orkgp}, and \textit{orkgsh} for classes, predicates, and shapes respectively. Hence, the LLM does not need to resolve and remember the URI (Universal Resource Identifier) of the entities and predicates, rather simply learns from the example questions and generates correct SPARQL queries. Second, the performance is biased due to the number of empty answer sets. For instance, as Table~\ref{tab:count_nulls} depicts, the number of null gold answers in dev is 14. The number of null system answers on the dev data via the three-shot and five-shot methods is 23 and 25 respectively. All those fourteen questions with null gold answers in dev, also have null system answers in both settings. Among all the fourteen questions that have null values, only one is from the same question in both methods due to syntax error. Furthermore, as the bottom part of Table~\ref{tab:count_nulls} shows, the number of null system answers on the test data is 27 with three-shot and 28 with the five-shot methods. However, the test data gold answer set is not publicly available as of this writing. Thus, we are unable to run a full analysis. In conclusion, removing the questions that have null answer sets from the dev and test, can reveal the gap between our model and others that participated in the challenge.

\begin{table}
  \caption{Number of null gold answers in dev; and null system answers in three and five-shot answer sets on dev and test data.}
  \label{tab:count_nulls}
  \begin{tabular}{lllll}
    \toprule
   & data & answer set & total & due to syntax error out of total\\
    \midrule
    & & gold & 14 & 1 \\
   & & three-shot & 23 & 7 \\
   Null answers & dev  & five-shot & 25 & 7  \\
    & & both in dev and three-shot & 14 & 1\\
     & & both in dev and five-shot & 14 & 1\\
    \midrule
     & test & three-shot & 27 & 3\\
      & & five-shot & 28 & 4 \\
    \bottomrule
  \end{tabular}
\end{table}

\subsection{Error Analysis}\label{erroranalysis}

Since the gold answer of the final phase is not available, our error analysis is based on the dev set experiment. Generally, the errors are 1) syntactic errors: which occur due to missing and improper placement of \textit{closing bracket (`\}')}, \textit{period (`.')}, and \textit{semicolon (`;')}. As shown in Table~\ref{tab:count_nulls}, on dev data both three-shot and five-shot methods generate 7 SPARQLs that have null results due to syntax error. Besides, on test data, the three-shot method generates 3 SPARQLs that have null answers due to syntactic errors. Likewise, the five-shot method has four SPARQL that are syntactically incorrect. 2) keyword matching: the entities identified by the LLM have extra or missing blank space(s). For example, the filter in ``\textit{SELECT ... FILTER (str(?dataset\_lbl) =` Jacquard dataset')...}" has a space at the beginning of the keyword `Jacquard dataset' in the gold SPARQL, but the LLM-generated SPARQL query does not have space. Thus, our model's SPARQL query results in a null answer. 3) Lack of question understanding: for complex questions like `Where can all the data sets used in the compared studies be found?', the LLM assumes that the question is about a dataset and generates ``\textit{SELECT DISTINCT ?dataset ?dataset\_lbl WHERE ...}" which looks for a dataset. However, the question is about the URI where the dataset is stored. Moreover, on questions like `What is the top benchmark score and its metric on the Words in Context dataset?', the LLM creates queries that look for the model and its name ``\textit{SELECT DISTINCT ?model ?model\_lbl WHERE ...}". Here the LLM misses the aim of the question, i.e., the correct answer is the top benchmark score and the respective query should look like ``\textit{SELECT DISTINCT ?metric ?metric\_lbl (MAX(?value) AS ?score) WHERE...}". Therefore, the misunderstanding of a question leads the LLM to produce SPARQLs that look for incorrect objects and miss operators like \textit{MAX()} in the recent example.

\section{Summary}\label{summary}

Our Scholarly KGQA system to the Scholarly-QALD-23 challenge at the ISWC 2023, follows a pipeline structure. For a given test question, the question analyzer identifies similar questions using a BERT-sentence encoder. Then, the SPARQL generator creates prompts by composing the top five (three or one) similar question-SPARQL pairs from the training set with the test question and generates a SPARQL by prompting Vicuna. Finally, the answer generator runs the query against the ORKG SPARQL endpoint and returns an answer. Our system achieves an F1-score of 99.0\% on the SciQA test set, which is a runner-up of the SciQA leaderboard\footnote{\url{https://kgqa.github.io/scholarly-QALD-challenge/2023/}}.

\begin{acknowledgments}
This work has been partially supported by grants for the DFG project NFDI4DataScience project (DFG project no. 460234259) and by the Federal Ministry for Economics and Climate Action in the project CoyPu (project number 01MK21007G). 
\end{acknowledgments}
%%
%% Define the bibliography file to be used
\bibliography{mybibliography}

\begin{thebibliography}{19}
\expandafter\ifx\csname natexlab\endcsname\relax\def\natexlab#1{#1}\fi
\providecommand{\url}[1]{\texttt{#1}}
\providecommand{\href}[2]{#2}
\providecommand{\path}[1]{#1}
\providecommand{\DOIprefix}{doi:}
\providecommand{\ArXivprefix}{arXiv:}
\providecommand{\URLprefix}{URL: }
\providecommand{\Pubmedprefix}{pmid:}
\providecommand{\doi}[1]{\href{http://dx.doi.org/#1}{\path{#1}}}
\providecommand{\Pubmed}[1]{\href{pmid:#1}{\path{#1}}}
\providecommand{\bibinfo}[2]{#2}
\ifx\xfnm\relax \def\xfnm[#1]{\unskip,\space#1}\fi
%Type = Article
\bibitem[{Auer et~al.(2020)Auer, Oelen, Haris, Stocker, D’Souza, Farfar, Vogt, Prinz, Wiens, and Jaradeh}]{auer2020improving}
\bibinfo{author}{S.~Auer}, \bibinfo{author}{A.~Oelen}, \bibinfo{author}{M.~Haris}, \bibinfo{author}{M.~Stocker}, \bibinfo{author}{J.~D’Souza}, \bibinfo{author}{K.~E. Farfar}, \bibinfo{author}{L.~Vogt}, \bibinfo{author}{M.~Prinz}, \bibinfo{author}{V.~Wiens}, \bibinfo{author}{M.~Y. Jaradeh},
\newblock \bibinfo{title}{{Improving access to scientific literature with knowledge graphs}},
\newblock \bibinfo{journal}{Bibliothek Forschung und Praxis} \bibinfo{volume}{44} (\bibinfo{year}{2020}) \bibinfo{pages}{516--529}. \URLprefix \url{https://www.degruyter.com/document/doi/10.1515/bfp-2020-2042/html}.
%Type = Inproceedings
\bibitem[{Jaradeh et~al.(2020)Jaradeh, Stocker, and Auer}]{jaradeh2020question}
\bibinfo{author}{M.~Y. Jaradeh}, \bibinfo{author}{M.~Stocker}, \bibinfo{author}{S.~Auer},
\newblock \bibinfo{title}{{Question Answering on Scholarly Knowledge Graphs}},
\newblock in: \bibinfo{booktitle}{International Conference on Theory and Practice of Digital Libraries}, \bibinfo{organization}{Springer}, \bibinfo{year}{2020}, pp. \bibinfo{pages}{19--32}. \URLprefix \url{https://link.springer.com/chapter/10.1007/978-3-030-54956-5_2}.
%Type = Inproceedings
\bibitem[{Wang et~al.(2021)Wang, Zhou, Zhang, and Wang}]{wang2021literatureqa}
\bibinfo{author}{H.~Wang}, \bibinfo{author}{L.~Zhou}, \bibinfo{author}{W.~Zhang}, \bibinfo{author}{X.~Wang},
\newblock \bibinfo{title}{{LiteratureQA: A Question Answering Corpus with Graph Knowledge on Academic Literature}},
\newblock in: \bibinfo{booktitle}{Proceedings of the 30th ACM International Conference on Information \& Knowledge Management}, \bibinfo{year}{2021}, pp. \bibinfo{pages}{4623--4632}. \URLprefix \url{https://dl.acm.org/doi/10.1145/3459637.3482007}.
%Type = Article
\bibitem[{Banerjee et~al.(2023)Banerjee, Awale, Usbeck, and Biemann}]{banerjee2023dblp}
\bibinfo{author}{D.~Banerjee}, \bibinfo{author}{S.~Awale}, \bibinfo{author}{R.~Usbeck}, \bibinfo{author}{C.~Biemann},
\newblock \bibinfo{title}{{DBLP-QuAD: A Question Answering Dataset over the DBLP Scholarly Knowledge Graph}},
\newblock \bibinfo{journal}{arXiv preprint arXiv:2303.13351}  (\bibinfo{year}{2023}). \URLprefix \url{http://arxiv.org/abs/2303.13351}.
%Type = Inproceedings
\bibitem[{Brown et~al.(2020)Brown, Mann, Ryder, Subbiah, Kaplan, Dhariwal, Neelakantan, Shyam, Sastry, Askell, Agarwal, Herbert-Voss, Krueger, Henighan, Child, Ramesh, Ziegler, Wu, Winter, Hesse, Chen, Sigler, Litwin, Gray, Chess, Clark, Berner, McCandlish, Radford, Sutskever, and Amodei}]{NEURIPS2020_1457c0d6}
\bibinfo{author}{T.~Brown}, \bibinfo{author}{B.~Mann}, \bibinfo{author}{N.~Ryder}, \bibinfo{author}{M.~Subbiah}, \bibinfo{author}{J.~D. Kaplan}, \bibinfo{author}{P.~Dhariwal}, \bibinfo{author}{A.~Neelakantan}, \bibinfo{author}{P.~Shyam}, \bibinfo{author}{G.~Sastry}, \bibinfo{author}{A.~Askell}, \bibinfo{author}{S.~Agarwal}, \bibinfo{author}{A.~Herbert-Voss}, \bibinfo{author}{G.~Krueger}, \bibinfo{author}{T.~Henighan}, \bibinfo{author}{R.~Child}, \bibinfo{author}{A.~Ramesh}, \bibinfo{author}{D.~Ziegler}, \bibinfo{author}{J.~Wu}, \bibinfo{author}{C.~Winter}, \bibinfo{author}{C.~Hesse}, \bibinfo{author}{M.~Chen}, \bibinfo{author}{E.~Sigler}, \bibinfo{author}{M.~Litwin}, \bibinfo{author}{S.~Gray}, \bibinfo{author}{B.~Chess}, \bibinfo{author}{J.~Clark}, \bibinfo{author}{C.~Berner}, \bibinfo{author}{S.~McCandlish}, \bibinfo{author}{A.~Radford}, \bibinfo{author}{I.~Sutskever}, \bibinfo{author}{D.~Amodei},
\newblock \bibinfo{title}{{Language Models are Few-Shot Learners}},
\newblock in: \bibinfo{editor}{H.~Larochelle}, \bibinfo{editor}{M.~Ranzato}, \bibinfo{editor}{R.~Hadsell}, \bibinfo{editor}{M.~Balcan}, \bibinfo{editor}{H.~Lin} (Eds.), \bibinfo{booktitle}{Advances in Neural Information Processing Systems}, volume~\bibinfo{volume}{33}, \bibinfo{publisher}{Curran Associates, Inc.}, \bibinfo{year}{2020}, pp. \bibinfo{pages}{1877--1901}. \URLprefix \url{https://proceedings.neurips.cc/paper_files/paper/2020/file/1457c0d6bfcb4967418bfb8ac142f64a-Paper.pdf}.
%Type = Article
\bibitem[{Touvron et~al.(2023)Touvron, Lavril, Izacard, Martinet, Lachaux, Lacroix, Rozi{\`e}re, Goyal, Hambro, Azhar et~al.}]{touvron2023llama}
\bibinfo{author}{H.~Touvron}, \bibinfo{author}{T.~Lavril}, \bibinfo{author}{G.~Izacard}, \bibinfo{author}{X.~Martinet}, \bibinfo{author}{M.-A. Lachaux}, \bibinfo{author}{T.~Lacroix}, \bibinfo{author}{B.~Rozi{\`e}re}, \bibinfo{author}{N.~Goyal}, \bibinfo{author}{E.~Hambro}, \bibinfo{author}{F.~Azhar}, et~al.,
\newblock \bibinfo{title}{{LLaMA: Open and efficient foundation language models}}  (\bibinfo{year}{2023}). \URLprefix \url{arXiv preprint arXiv:2302.13971}.
%Type = Inproceedings
\bibitem[{Banerjee et~al.(2023)Banerjee, Nair, Usbeck, and Biemann}]{banerjee2023gett}
\bibinfo{author}{D.~Banerjee}, \bibinfo{author}{P.~A. Nair}, \bibinfo{author}{R.~Usbeck}, \bibinfo{author}{C.~Biemann},
\newblock \bibinfo{title}{{GETT-QA: Graph Embedding Based T2T Transformer for Knowledge Graph Question Answering}},
\newblock in: \bibinfo{booktitle}{European Semantic Web Conference}, \bibinfo{organization}{Springer}, \bibinfo{year}{2023}, pp. \bibinfo{pages}{279--297}. \URLprefix \url{https://link.springer.com/content/pdf/10.1007/978-3-031-33455-9_17.pdf}.
%Type = Inproceedings
\bibitem[{Chen(2023)}]{chen-2023-large}
\bibinfo{author}{W.~Chen},
\newblock \bibinfo{title}{{Large Language Models are few(1)-shot Table Reasoners}},
\newblock in: \bibinfo{booktitle}{Findings of the Association for Computational Linguistics: EACL 2023}, \bibinfo{publisher}{Association for Computational Linguistics}, \bibinfo{address}{Dubrovnik, Croatia}, \bibinfo{year}{2023}, pp. \bibinfo{pages}{1120--1130}. \URLprefix \url{https://aclanthology.org/2023.findings-eacl.83}. \DOIprefix\doi{10.18653/v1/2023.findings-eacl.83}.
%Type = Inproceedings
\bibitem[{Kamalloo et~al.(2023)Kamalloo, Dziri, Clarke, and Rafiei}]{kamalloo-etal-2023-evaluating}
\bibinfo{author}{E.~Kamalloo}, \bibinfo{author}{N.~Dziri}, \bibinfo{author}{C.~Clarke}, \bibinfo{author}{D.~Rafiei},
\newblock \bibinfo{title}{{Evaluating Open-Domain Question Answering in the Era of Large Language Models}},
\newblock in: \bibinfo{booktitle}{Proceedings of the 61st Annual Meeting of the Association for Computational Linguistics (Volume 1: Long Papers)}, \bibinfo{publisher}{Association for Computational Linguistics}, \bibinfo{address}{Toronto, Canada}, \bibinfo{year}{2023}, pp. \bibinfo{pages}{5591--5606}. \URLprefix \url{https://aclanthology.org/2023.acl-long.307}. \DOIprefix\doi{10.18653/v1/2023.acl-long.307}.
%Type = Inproceedings
\bibitem[{Ziems et~al.(2023)Ziems, Yu, Zhang, and Jiang}]{ziems-etal-2023-large}
\bibinfo{author}{N.~Ziems}, \bibinfo{author}{W.~Yu}, \bibinfo{author}{Z.~Zhang}, \bibinfo{author}{M.~Jiang},
\newblock \bibinfo{title}{{Large Language Models are Built-in Autoregressive Search Engines}},
\newblock in: \bibinfo{booktitle}{Findings of the Association for Computational Linguistics: ACL 2023}, \bibinfo{publisher}{Association for Computational Linguistics}, \bibinfo{address}{Toronto, Canada}, \bibinfo{year}{2023}, pp. \bibinfo{pages}{2666--2678}. \URLprefix \url{https://aclanthology.org/2023.findings-acl.167}. \DOIprefix\doi{10.18653/v1/2023.findings-acl.167}.
%Type = Inproceedings
\bibitem[{Sorensen et~al.(2022)Sorensen, Robinson, Rytting, Shaw, Rogers, Delorey, Khalil, Fulda, and Wingate}]{sorensen-etal-2022-information}
\bibinfo{author}{T.~Sorensen}, \bibinfo{author}{J.~Robinson}, \bibinfo{author}{C.~Rytting}, \bibinfo{author}{A.~Shaw}, \bibinfo{author}{K.~Rogers}, \bibinfo{author}{A.~Delorey}, \bibinfo{author}{M.~Khalil}, \bibinfo{author}{N.~Fulda}, \bibinfo{author}{D.~Wingate},
\newblock \bibinfo{title}{{An Information-theoretic Approach to Prompt Engineering Without Ground Truth Labels}},
\newblock in: \bibinfo{booktitle}{Proceedings of the 60th Annual Meeting of the Association for Computational Linguistics (Volume 1: Long Papers)}, \bibinfo{publisher}{Association for Computational Linguistics}, \bibinfo{address}{Dublin, Ireland}, \bibinfo{year}{2022}, pp. \bibinfo{pages}{819--862}. \URLprefix \url{https://aclanthology.org/2022.acl-long.60}. \DOIprefix\doi{10.18653/v1/2022.acl-long.60}.
%Type = Article
\bibitem[{Kocaman and Talby(2021)}]{KOCAMAN2021100058}
\bibinfo{author}{V.~Kocaman}, \bibinfo{author}{D.~Talby},
\newblock \bibinfo{title}{{Spark NLP: Natural language understanding at scale}},
\newblock \bibinfo{journal}{Software Impacts}  (\bibinfo{year}{2021}) \bibinfo{pages}{100058}. \URLprefix \url{https://www.sciencedirect.com/science/article/pii/S2665963821000063}. \DOIprefix\doi{https://doi.org/10.1016/j.simpa.2021.100058}.
%Type = Article
\bibitem[{Zhang et~al.(2023)Zhang, Zhang, Ke, Li, Huang, Shao, Cao, and Lv}]{ZHANG20231}
\bibinfo{author}{L.~Zhang}, \bibinfo{author}{J.~Zhang}, \bibinfo{author}{X.~Ke}, \bibinfo{author}{H.~Li}, \bibinfo{author}{X.~Huang}, \bibinfo{author}{Z.~Shao}, \bibinfo{author}{S.~Cao}, \bibinfo{author}{X.~Lv},
\newblock \bibinfo{title}{{A survey on complex factual question answering}},
\newblock \bibinfo{journal}{AI Open} \bibinfo{volume}{4} (\bibinfo{year}{2023}) \bibinfo{pages}{1--12}. \URLprefix \url{https://www.sciencedirect.com/science/article/pii/S2666651022000249}. \DOIprefix\doi{https://doi.org/10.1016/j.aiopen.2022.12.003}.
%Type = Inproceedings
\bibitem[{Devlin et~al.(2019)Devlin, Chang, Lee, and Toutanova}]{bertDevlinCLT19}
\bibinfo{author}{J.~Devlin}, \bibinfo{author}{M.~Chang}, \bibinfo{author}{K.~Lee}, \bibinfo{author}{K.~Toutanova},
\newblock \bibinfo{title}{{{BERT:} Pre-training of Deep Bidirectional Transformers for Language Understanding}},
\newblock in: \bibinfo{editor}{J.~Burstein}, \bibinfo{editor}{C.~Doran}, \bibinfo{editor}{T.~Solorio} (Eds.), \bibinfo{booktitle}{Proceedings of the 2019 Conference of the North American Chapter of the Association for Computational Linguistics: Human Language Technologies, {NAACL-HLT} 2019, Minneapolis, MN, USA, June 2-7, 2019, Volume 1 (Long and Short Papers)}, \bibinfo{publisher}{Association for Computational Linguistics}, \bibinfo{year}{2019}, pp. \bibinfo{pages}{4171--4186}. \URLprefix \url{https://doi.org/10.18653/v1/n19-1423}. \DOIprefix\doi{10.18653/v1/n19-1423}.
%Type = Article
\bibitem[{Raffel et~al.(2020)Raffel, Shazeer, Roberts, Lee, Narang, Matena, Zhou, Li, and Liu}]{t5raffel}
\bibinfo{author}{C.~Raffel}, \bibinfo{author}{N.~Shazeer}, \bibinfo{author}{A.~Roberts}, \bibinfo{author}{K.~Lee}, \bibinfo{author}{S.~Narang}, \bibinfo{author}{M.~Matena}, \bibinfo{author}{Y.~Zhou}, \bibinfo{author}{W.~Li}, \bibinfo{author}{P.~J. Liu},
\newblock \bibinfo{title}{{Exploring the Limits of Transfer Learning with a Unified Text-to-Text Transformer}},
\newblock \bibinfo{journal}{J. Mach. Learn. Res.} \bibinfo{volume}{21} (\bibinfo{year}{2020}). \URLprefix \url{https://jmlr.org/papers/volume21/20-074/20-074.pdf}.
%Type = Inproceedings
\bibitem[{Baek et~al.(2023)Baek, Aji, and Saffari}]{baek-etal-2023-knowledge}
\bibinfo{author}{J.~Baek}, \bibinfo{author}{A.~F. Aji}, \bibinfo{author}{A.~Saffari},
\newblock \bibinfo{title}{{Knowledge-Augmented Language Model Prompting for Zero-Shot Knowledge Graph Question Answering}},
\newblock in: \bibinfo{booktitle}{Proceedings of the 1st Workshop on Natural Language Reasoning and Structured Explanations (NLRSE)}, \bibinfo{publisher}{Association for Computational Linguistics}, \bibinfo{address}{Toronto, Canada}, \bibinfo{year}{2023}, pp. \bibinfo{pages}{78--106}. \URLprefix \url{https://aclanthology.org/2023.nlrse-1.7}. \DOIprefix\doi{10.18653/v1/2023.nlrse-1.7}.
%Type = Article
\bibitem[{Usbeck et~al.(2023)Usbeck, Yan, Perevalov, Jiang, Schulz, Kraft, M{\"o}ller, Huang, Reineke, Ngomo et~al.}]{usbeckqald}
\bibinfo{author}{R.~Usbeck}, \bibinfo{author}{X.~Yan}, \bibinfo{author}{A.~Perevalov}, \bibinfo{author}{L.~Jiang}, \bibinfo{author}{J.~Schulz}, \bibinfo{author}{A.~Kraft}, \bibinfo{author}{C.~M{\"o}ller}, \bibinfo{author}{J.~Huang}, \bibinfo{author}{J.~Reineke}, \bibinfo{author}{A.-C.~N. Ngomo}, et~al.,
\newblock \bibinfo{title}{{QALD-10—The 10th Challenge on Question Answering over Linked Data}},
\newblock \bibinfo{journal}{Semantic Web}  (\bibinfo{year}{2023}). \URLprefix \url{https://www.semantic-web-journal.net/system/files/swj3471.pdf}.
%Type = Article
\bibitem[{Auer et~al.(2023)Auer, Barone, Bartz, Cortes, Jaradeh, Karras, Koubarakis, Mouromtsev, Pliukhin, Radyush et~al.}]{auer2023sciqa}
\bibinfo{author}{S.~Auer}, \bibinfo{author}{D.~A. Barone}, \bibinfo{author}{C.~Bartz}, \bibinfo{author}{E.~G. Cortes}, \bibinfo{author}{M.~Y. Jaradeh}, \bibinfo{author}{O.~Karras}, \bibinfo{author}{M.~Koubarakis}, \bibinfo{author}{D.~Mouromtsev}, \bibinfo{author}{D.~Pliukhin}, \bibinfo{author}{D.~Radyush}, et~al.,
\newblock \bibinfo{title}{{The SciQA Scientific Question Answering Benchmark for Scholarly Knowledge}},
\newblock \bibinfo{journal}{Scientific Reports} \bibinfo{volume}{13} (\bibinfo{year}{2023}) \bibinfo{pages}{7240}. \URLprefix \url{https://www.nature.com/articles/s41598-023-33607-z}.
%Type = Article
\bibitem[{Pavao et~al.(2023)Pavao, Guyon, Letournel, Tran, Baro, Escalante, Escalera, Thomas, and Xu}]{codalab_competitions_JMLR}
\bibinfo{author}{A.~Pavao}, \bibinfo{author}{I.~Guyon}, \bibinfo{author}{A.-C. Letournel}, \bibinfo{author}{D.-T. Tran}, \bibinfo{author}{X.~Baro}, \bibinfo{author}{H.~J. Escalante}, \bibinfo{author}{S.~Escalera}, \bibinfo{author}{T.~Thomas}, \bibinfo{author}{Z.~Xu},
\newblock \bibinfo{title}{{CodaLab Competitions: An Open Source Platform to Organize Scientific Challenges}},
\newblock \bibinfo{journal}{Journal of Machine Learning Research} \bibinfo{volume}{24} (\bibinfo{year}{2023}) \bibinfo{pages}{1--6}. \URLprefix \url{http://jmlr.org/papers/v24/21-1436.html}.

\end{thebibliography}

\end{document}